\pdfoutput=1

\documentclass{article}
\usepackage[T1]{fontenc}

\usepackage{amsmath,amssymb,amsthm}

\usepackage[
  paper  = letterpaper,
  left   = 1.5in,
  right  = 1.5in,
  top    = 1.0in,
  bottom = 1.0in,
  ]{geometry}

\usepackage[usenames,dvipsnames]{xcolor}
\usepackage{graphicx}
\usepackage[labelfont=bf]{caption}
\usepackage[format=hang]{subcaption}
\usepackage{booktabs}
\usepackage{nicefrac}
\usepackage[numbers,sort]{natbib}
\usepackage[algoruled]{algorithm2e}
\usepackage[acronym,smallcaps,nowarn,nomain]{glossaries}
\makeglossaries

\usepackage{listings}
\lstdefinestyle{mystyle}{
    commentstyle=\color{OliveGreen},
    numberstyle=\tiny\color{black!60},
    stringstyle=\color{BrickRed},
    basicstyle=\ttfamily\scriptsize,
    breakatwhitespace=false,
    breaklines=true,
    captionpos=b,
    keepspaces=true,
    numbers=none,
    numbersep=5pt,
    showspaces=false,
    showstringspaces=false,
    showtabs=false,
    tabsize=2
}
\DeclareRobustCommand{\parhead}[1]{\textbf{#1}}

\newcommand{\KL}[2]{\ensuremath{\textrm{KL}\left(#1\;\|\;#2\right)}}

\DeclareMathOperator*{\argmax}{arg\,max}

\DeclareRobustCommand{\mb}[1]{\boldsymbol{\mathbf{#1}}}

\newcommand{\mbx}{\mb{x}}
\newcommand{\mbX}{\mb{X}}

\newcommand{\mbI}{\mb{I}}

\newcommand{\mbtheta}{{\theta}}

\newcommand{\mbomega}{{\omega}}

\newcommand{\mbsigma}{{\sigma}}

\newcommand{\mbzeta}{{\zeta}}
\newcommand{\mbeta}{{\eta}}
\newcommand{\mbbeta}{{\beta}}
\newcommand{\mbphi}{{\phi}}
\newcommand{\mbmu}{{\mu}}
\newcommand{\mbrho}{{\rho}}

\newcommand\dif{\mathop{}\!\mathrm{d}}
\newcommand{\diag}{\textrm{diag}}
\newcommand{\supp}{\textrm{supp}}

\newcommand{\E}{\mathbb{E}}

\newcommand{\bbR}{\mathbb{R}}

\newcommand{\cL}{\mathcal{L}}

\newcommand{\cN}{\mathcal{N}}
\newcommand{\Gam}{\textrm{Gam}}

\newacronym{KL}{kl}{Kullback-Leibler}
\newacronym{ELBO}{elbo}{evidence lower bound}

\newacronym{MC}{mc}{Monte Carlo}
\newacronym{MCMC}{mcmc}{Markov chain Monte Carlo}

\newacronym{VI}{vi}{variational inference}
\newacronym{MFVI}{mfvi}{mean-field variational inference}
\newacronym{SVI}{svi}{stochastic variational inference}

\newacronym{VMP}{vmp}{variational message passing}

\newacronym{ADVI}{advi}{automatic differentiation variational inference}

\newacronym{RMSPROP}{rmsprop}{rmsprop}

\newacronym{NUTS}{nuts}{no-U-turn sampler}
\newacronym{HMC}{hmc}{Hamiltonian Monte Carlo}

\newacronym{ARD}{ard}{automatic relevance determination}
\newacronym{GMM}{gmm}{Gaussian mixture model}

\definecolor{POSTcolor}{rgb}{0.48, 0.20, 0.58}
\definecolor{Qcolor}{rgb}{0.00, 0.53, 0.22}

\title{\textbf{Automatic Variational Inference in Stan}}

\author{%
\textbf{Alp Kucukelbir}                              \\
Data Science Institute                      \\
Department of Computer Science              \\
Columbia University                         \\
\texttt{alp@cs.columbia.edu}                \\\\
\textbf{Rajesh Ranganath}                            \\
Department of Computer Science              \\
Princeton University                        \\
\texttt{rajeshr@cs.princeton.edu}           \\\\
\textbf{Andrew Gelman}                               \\
Data Science Institute                      \\
Depts.~of Political Science, Statistics     \\
Columbia University                         \\
\texttt{gelman@stat.columbia.edu}           \\\\
\textbf{David M.~Blei}                               \\
Data Science Institute                      \\
Depts.~of Computer Science, Statistics      \\
Columbia University                         \\
\texttt{david.blei@columbia.edu}
}

\begin{document}
\maketitle

\begin{abstract}
Variational inference is a scalable technique for approximate Bayesian
inference.
Deriving variational inference algorithms requires tedious model-specific
calculations; this makes it difficult to automate.
We propose an automatic variational inference algorithm, \gls{ADVI}.
The user only provides a Bayesian model and a dataset; nothing else.
We make no conjugacy assumptions and support a broad class of models.
The algorithm automatically determines an appropriate variational family and
optimizes the variational objective.
We implement \gls{ADVI} in Stan (code available now), a probabilistic
programming framework. We compare \gls{ADVI} to \textsc{mcmc} sampling across
hierarchical generalized linear models, nonconjugate matrix factorization,
and a mixture model. We train the mixture model on a quarter million images.
With \gls{ADVI} we can use variational inference on any
model we write in Stan.
\end{abstract}

\glsresetall{}
\section{Introduction}

Bayesian inference is a powerful framework for analyzing data. We design a
model for data using latent variables; we then analyze
data by calculating the posterior density of the latent variables. For
machine learning models, calculating the posterior is often difficult; we resort
to approximation.

\Gls{VI} approximates the posterior with a simpler density
\citep{jordan1999introduction,wainwright2008graphical}.
We search over a family of simple densities and find the member
closest to the posterior. This turns approximate
inference into optimization.  \gls{VI} has had a tremendous impact on
machine learning; it is typically faster than \gls{MCMC}
sampling (as we show here too) and has recently scaled up to
massive data~\citep{hoffman2013stochastic}.

Unfortunately, \gls{VI} algorithms are difficult to
derive. We must first define the family of approximating
densities, and then calculate model-specific quantities relative
to that family to solve the variational optimization problem.  Both
steps require expert knowledge.  The resulting algorithm is tied to
both the model and the chosen approximation.

In this paper we develop a method for automating variational
inference, \gls{ADVI}.  Given any model from a wide class
(specifically, differentiable probability models), \gls{ADVI}
determines an appropriate variational family and an algorithm for
optimizing the corresponding variational objective.  We implement
\gls{ADVI} in Stan \citep{stan-manual:2015}, a flexible probabilistic
programming framework originally designed for sampling-based
inference.  Stan describes a high-level language to define
probabilistic models (e.g., Figure \ref{fig:example_poisson}) as well as a
model compiler, a library of transformations, and an efficient
automatic differentiation toolbox.  With \gls{ADVI} we can now use
variational inference on any model we can express in Stan.%
\footnote{\gls{ADVI} is available in Stan 2.7 (development branch). It will
appear in Stan 2.8. See Appendix \ref{app:stan}.}
(See Appendices \ref{app:linreg_ard} to \ref{app:gmm}.)

\begin{figure}[!t]
\centering
  \begin{subfigure}[t]{2.6in}
    \includegraphics[width=2.6in]{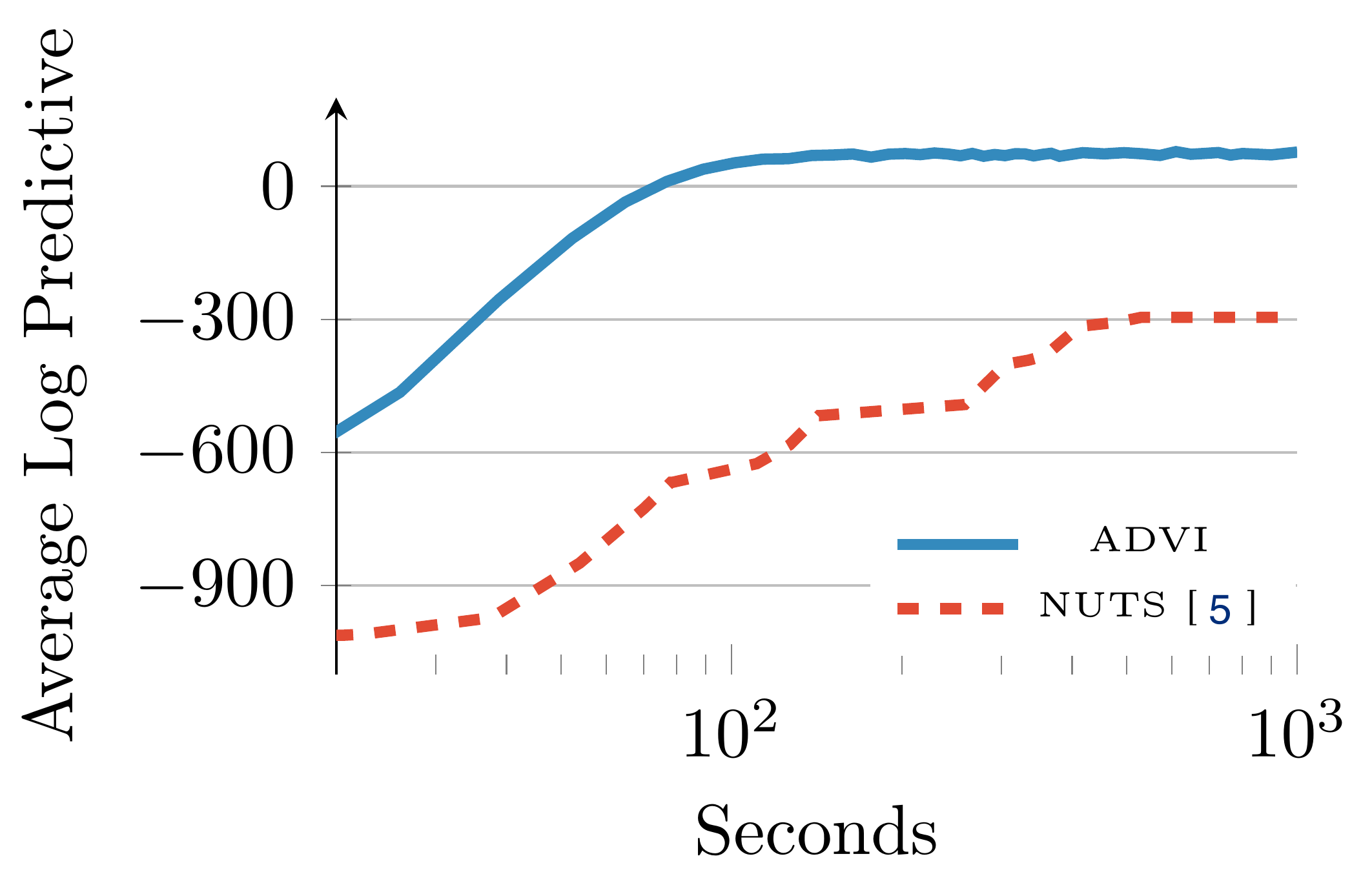}
    \vspace*{-0.12in}
    \caption{Subset of $1000$ images}
    \label{sub:gmm_1000}
  \end{subfigure}
  \hspace*{0.15in}
  \begin{subfigure}[t]{2.6in}
    \includegraphics[width=2.6in]{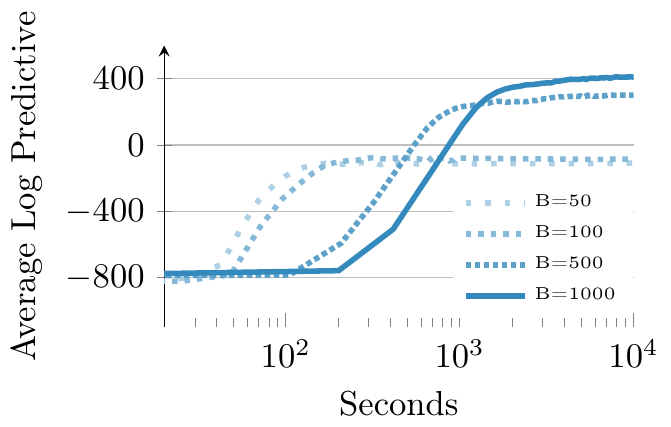}
    \vspace*{-0.12in}
    \caption{Full dataset of $250\,000$ images}
    \label{sub:gmm_adsvi}
  \end{subfigure}
  \caption{
  Held-out predictive accuracy results | \gls{GMM} of the image\textsc
  {clef} image histogram dataset. \textbf{(a)} \gls{ADVI}
  outperforms the \gls {NUTS}, the default sampling method in Stan \citep
  {hoffman2014nuts}.
  \textbf{(b)} \gls{ADVI} scales to large datasets by subsampling minibatches
  of size $B$ from the dataset at each iteration \citep{hoffman2013stochastic}.
  We present more details in Section \ref{sub:scaling_gmm} and Appendix
  \ref{app:gmm}.
  }
  \label{fig:gmm_plots}
\end{figure}

Figure \ref{fig:gmm_plots} illustrates the advantages of our method.  We
present a nonconjugate Gaussian mixture model for analyzing natural
images; this is 40 lines in Stan (Figure \ref{fig:code_gmm_diag}).
Section \ref{sub:gmm_1000}
illustrates Bayesian inference on $1000$ images.  The $y$-axis is held-out
likelihood, a measure of model fitness; the $x$-axis is time (on a
log scale). \gls{ADVI} is orders of magnitude faster than \gls{NUTS},
a state-of-the-art \gls{MCMC} algorithm (and Stan's default inference
technique) \citep{hoffman2014nuts}.
We also study nonconjugate factorization models and
hierarchical generalized linear models; we consistently observe
speed-up against \gls{NUTS}.

Section \ref{sub:gmm_adsvi} illustrates Bayesian inference on $250\,000$ images,
the size of data we more commonly find in machine learning.  Here we use
\gls{ADVI} with stochastic variational
inference~\citep{hoffman2013stochastic}, giving an approximate
posterior in under two hours.  For data like these, \gls{MCMC} techniques cannot
even practically begin analysis, a motivating case for approximate
inference.

\parhead{Related Work.} \gls{ADVI} automates variational
inference within the Stan probabilistic programming framework
\citep{stan-manual:2015}. This draws on two major themes.

The first is a body of work that aims to generalize \gls{VI}. \citet
{kingma2013auto}
and \citet{rezende2014stochastic} describe a reparameterization of the
variational problem that simplifies optimization.
\citet{ranganath2014black} and \citet{salimans2014using} propose a
black-box technique that only uses the gradient of the approximating family for
optimization. \citet{titsias2014doubly} leverage the gradient of the model for a
small class of models.  We build on and extend these ideas to automate
variational inference; we highlight technical connections as we develop our
method.

The second theme is probabilistic
programming. \citet{wingate2013automated} study \gls{VI} in general
probabilistic programs, as supported by languages like Church
\citep{goodman2008church}, Venture \citep{mansinghka2014venture}, and
Anglican \citep{wood2014new}. Another probabilistic programming
framework is infer.NET, which implements variational message passing
\citep {winn2005variational}, an efficient algorithm for conditionally
conjugate graphical models. Stan supports a more comprehensive class of models
that we describe in Section \ref{sub:diff_prob_models}.

\glsresetall{}
\section{Automatic Differentiation Variational Inference}

\Gls{ADVI} follows a straightforward recipe. First, we
transform the
space of the latent variables in our model to the real coordinate
space. For example, the logarithm transforms a positively constrained
variable, such as a standard deviation, to the real line. Then, we
posit a Gaussian variational distribution. This induces a
non-Gaussian approximation in the original variable space. Last, we combine
automatic differentiation with stochastic optimization to maximize the
variational objective. We begin by defining the class of models we support.

\subsection{Differentiable Probability Models}
\label{sub:diff_prob_models}

Consider a dataset $\mbX = \mbx_{1:N}$ with $N$ observations. Each $\mbx_n$ is a
discrete or continuous random vector. The
likelihood $p (\mbX\mid\mbtheta)$ relates the observations to a set of
latent random variables $\mbtheta$. Bayesian analysis posits a prior
density $p(\mbtheta)$ on the latent variables. Combining the
likelihood with the prior gives the joint density
$p(\mbX,\mbtheta) = p(\mbX\mid\mbtheta)\,p(\mbtheta)$.

We focus on approximate inference for differentiable probability models.
These models have continuous latent variables $\mbtheta$. They also have a
gradient of the log-joint with respect to the latent variables $\nabla_\mbtheta
\log p(\mbX,\mbtheta)$. The gradient is valid within the support of the prior
$ \supp(p(\mbtheta)) = \big\{\, \mbtheta \mid \mbtheta \in \bbR^K
\text{ and } p(\mbtheta) > 0 \,\big\} \subseteq \bbR^{K}$,
where $K$ is the dimension of the latent variable space. This support set is
important: it determines the support of the posterior density and will play
an important role later in the paper. Note that we make no assumptions about
conjugacy, either full%
\footnote{The posterior of a \emph{fully} conjugate model is in the same family
as the prior.}
or conditional.%
\footnote{A \emph{conditionally} conjugate model has this property
within the complete conditionals of the model \citep{hoffman2013stochastic}.}

\begin{figure}[tbp]
\centering
    \parbox[c][1.70in]{0.75in}{%
    \vspace*{-0.15in}
    \includegraphics[width=0.5in]{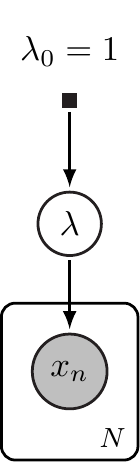}
    }
    \parbox[c][1.70in]{3.20in}{%
    \input{img/poisson_model_code.tex}
    }
  \caption{Specifying a simple nonconjugate probability model in Stan.}
  \label{fig:example_poisson}
\end{figure}

Consider a model that contains a Poisson likelihood with
unknown rate, $p(x\mid\lambda)$.  The observed variable $x$ is
discrete; the latent rate $\lambda$ is continuous and
positive.  Place an exponential prior for
$\lambda$, defined over the positive real numbers. The resulting joint
density describes a nonconjugate differentiable probability model. (See
Figure \ref{fig:example_poisson}.)  Its partial derivative
$\partial/\partial\lambda \, p (x,\lambda)$ is valid within the
support of the exponential distribution,
$\supp(p(\lambda)) = \bbR^+ \subset \bbR$. Since this model is nonconjugate,
the posterior is not an exponential distribution. This presents a challenge for
classical variational inference. We will see how \gls{ADVI} handles this model
later in the paper.

Many machine learning models are differentiable probability models. Linear and
logistic regression, matrix factorization with continuous or discrete
measurements, linear dynamical systems, and Gaussian processes are prime
examples. In machine learning, we usually describe mixture models, hidden Markov
models, and topic models with discrete random variables. Marginalizing out the
discrete variables reveals that these are also differentiable probability models.
(We show an example in Section \ref{sub:scaling_gmm}.) Only fully discrete
models,
such as the Ising model, fall outside of this category.

\glsreset{VI}
\subsection{Variational Inference}
\label{sub:variational_inference}

In Bayesian inference, we seek the posterior density
$p (\mbtheta\mid\mbX)$, which describes how the latent variables vary,
conditioned on a set of observations $\mbX$. Many posterior densities are
intractable because they lack analytic (closed-form) solutions. Thus, we seek to
approximate the posterior.

Consider an approximating density $q (\mbtheta\,;\,\mbphi)$
parameterized by $\mbphi$. We make no assumptions about its
shape or support. We want to find the
parameters of $q(\mbtheta\,;\,\mbphi)$ to best match the posterior
according to some loss function.  \Gls{VI} minimizes the \gls{KL}
divergence,
\begin{align}
  \min_\mbphi \KL{q (\mbtheta\,;\,\mbphi)}{p(\mbtheta\mid\mbX)},
  \label{eq:min_KL}
\end{align}
from the approximation to the posterior \citep{wainwright2008graphical}.
Typically the \gls{KL} divergence also lacks an analytic
form. Instead we maximize a proxy to the \gls{KL} divergence, the \gls{ELBO}
\begin{align*}
  \cL (\mbphi)
  &=
  \E_{q (\mbtheta)} \big[ \log p (\mbX,\mbtheta) \big]
  -
  \E_{q (\mbtheta)} \big[ \log q (\mbtheta\,;\,\mbphi) \big].
\end{align*}
The first term is an expectation of the
joint density under the approximation, and the second is the entropy of the
variational density. Maximizing the \gls{ELBO} minimizes the \gls{KL}
divergence \citep{jordan1999introduction,bishop2006pattern}.

The minimization problem from Equation \ref{eq:min_KL} becomes
\begin{align}
  \mbphi^*
  &=
  \argmax_\mbphi \cL(\mbphi)
  \quad\text{such that}\quad
  \supp(q(\mbtheta\,;\,\mbphi)) \subseteq \supp(p(\mbtheta \mid \mbX)),
  \label{eq:max_elbo}
\end{align}
where we explicitly specify the support matching constraint implied in the
\gls{KL} divergence.%
\footnote{If $\supp(q) \not\subseteq \supp (p)$
then outside the support of $p$ we have $\KL {q} {p} = \E_q [\log q] -
\E_q[\log p] = -\infty$.}
We highlight this constraint, as we do not specify the form of the
variational approximation; thus we must ensure that $q(\mbtheta\,;\,\mbphi)$
stays within the support of the posterior, which is equal to the support of the
prior.

\parhead{Why is \gls{VI} difficult to automate?}
In classical variational inference, we typically design a conditionally
conjugate model; the optimal approximating family matches the prior,
which satisfies the support constraint by definition \citep{bishop2006pattern}.
In other models, we carefully study the model and design custom
approximations. These depend on the model and on the choice of the
approximating density.

One way to automate \gls{VI} is to use black-box variational inference
\citep{ranganath2014black,salimans2014using}. If we select a density
whose support matches the posterior, then we can directly maximize the \gls
{ELBO} using \gls {MC} integration and stochastic optimization.
Another strategy is to restrict the class of models and use a fixed
variational approximation \citep{titsias2014doubly}. For instance, we may
use a Gaussian density for inference in unrestrained differentiable
probability models, i.e.~where $\supp (p (\mbtheta)) = \bbR^K$.

We adopt a transformation-based approach. First, we automatically
transform the support of the latent variables in our model to the real
coordinate space. Then, we posit a Gaussian variational density. The inverse of
our transform induces a  non-Gaussian variational approximation in the
original variable space. The transformation guarantees that the non-Gaussian
approximation stays within the support of the posterior. Here is how it works.

\subsection{Automatic Transformation of Constrained Variables}

Begin by transforming the support of the latent variables $\mbtheta$
such that they live in the real coordinate space $\bbR^K$.  Define a one-to-one
differentiable function
\begin{align}
  T &: \text{supp}(p(\mbtheta)) \rightarrow \bbR^K,
  \label{eq:transformation}
\end{align}
and identify the transformed variables as $\mbzeta = T(\mbtheta)$.
The transformed joint density $g(\mbX,\mbzeta)$ is a function of $\mbzeta$; it
has the
representation
\begin{align*}
  g(\mbX,\mbzeta)
  &=
  p
  (\mbX,T^{-1}(\mbzeta))
  \big|
  \det J_{T^{-1}}(\mbzeta)
  \big|,
\end{align*}
where $p$ is the joint density in the original latent variable space,
and $J_{T^{-1}}(\mbzeta)$ is the Jacobian of the
inverse of $T$. Transformations of continuous probability densities require a
Jacobian; it accounts for how the transformation warps unit volumes \citep
{olive2014statistical}. (See Appendix \ref{app:jacobian}.)

Consider again our running example.
The rate $\lambda$ lives in $\bbR^+$. The logarithm
$\zeta = T(\lambda) = \log(\lambda)$ transforms $\bbR^+$ to the real line
$\bbR$. Its
Jacobian adjustment is the derivative of the inverse of the logarithm,
$|\det J_{T^{-1}(\zeta)}| = \exp(\zeta)$.
The transformed density is
$
g(x,\zeta) =
\text{Poisson}(x\mid\exp(\zeta))
\,
\text{Exponential}(\exp(\zeta))
\,
\exp(\zeta)
$.
Figures \ref{sub:A} and \ref{sub:B} depict this transformation.

As we describe in the introduction, we implement our algorithm in Stan to enable
generic inference.
Stan implements a model compiler that automatically handles
transformations. It works by applying a library of
transformations and their corresponding Jacobians to the joint model density.%
\footnote{Stan
provides transformations for upper and lower bounds, simplex and ordered
vectors, and structured matrices such as covariance matrices and Cholesky
factors \citep{stan-manual:2015}.}
This transforms the joint density of any differentiable probability model to the
real coordinate space. Now, we can choose a variational distribution
independent from the model.

\begin{figure}[htb]
\input{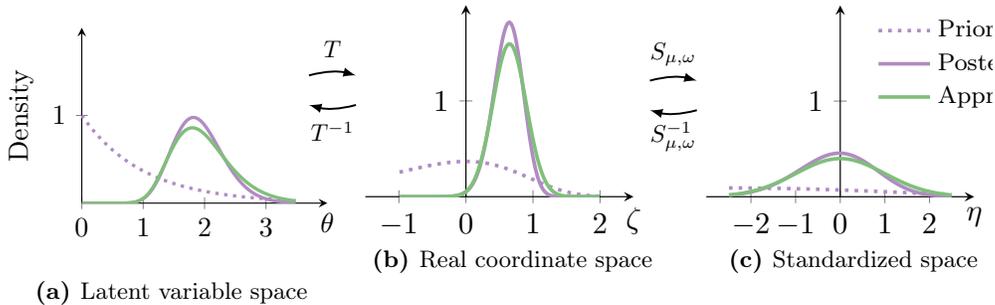}
\caption{Transformations for \gls{ADVI}. The \textcolor{POSTcolor}{purple} line
is the posterior. The \textcolor{Qcolor}{green} line is the approximation.
\textbf{(a)}
The latent variable space is $\bbR^+$.
\textbf{(a$\to$b)}
$T$ transforms the latent variable space to $\bbR$.
\textbf{(b)}
The variational approximation is a Gaussian.
\textbf{(b$\to$c)}
$S_{\mbmu,\mbomega}$ absorbs the parameters of the Gaussian.
\textbf{(c)}
We maximize the \gls{ELBO} in the standardized space, with a fixed
standard Gaussian approximation.}
\label{fig:transformations}
\end{figure}

\subsection{Implicit Non-Gaussian Variational Approximation}

After the transformation, the latent variables $\mbzeta$ have support
on $\bbR^K$. We posit a diagonal (mean-field) Gaussian variational
approximation
\begin{align*}
  q(\mbzeta \,;\, \mbphi)
  &=
  \cN(\mbzeta \,;\, \mbmu, \mbsigma^2)
  =
  \prod_{k=1}^K
  \cN
  (\zeta_k \,;\, \mu_k, \sigma^2_k),
\end{align*}
where the vector
$\mbphi = (\mu_{1},\cdots,\mu_{K}, \sigma^2_ {1},\cdots,\sigma^2_{K})$
concatenates the mean and variance of each Gaussian factor. This defines our
variational approximation in the real coordinate space. (Figure \ref{sub:B}.)

The transformation $T$ from Equation \ref{eq:transformation} maps the support of
the
latent variables to the real coordinate space. Thus, its inverse $T^{-1}$ maps
back to the support of the latent variables. This implicitly defines the
variational approximation in the original latent variable space as
$%
\cN (T^{-1}(\mbzeta)\,;\, \mbmu, \mbsigma^2)
\big| \det J_{T^{-1}}(\mbzeta) \big|.
$
The transformation ensures that the support of this approximation is always
bounded by that of the true posterior in the original latent variable space
(Figure \ref{sub:A}). Thus we can freely optimize the \gls{ELBO} in the
real coordinate space (Figure \ref{sub:B}) without worrying about the support
matching
constraint.

The \gls{ELBO} in the real coordinate space is
\begin{align*}
  \cL(\mbmu,\mbsigma^2)
  &=
  \E_{q(\mbzeta)}
  \bigg[
  \log p (\mbX, T^{-1}(\mbzeta))
  +
  \log \big| \det J_{T^{-1}}(\mbzeta) \big|
  \bigg]
  +
  \frac{K}{2}\left(1+\log(2\pi)\right)
  +
  \sum_{k=1}^K \log \sigma_k,
\end{align*}
where we plug in the analytic form for the Gaussian entropy. (Derivation in
Appendix \ref{app:elbo_unconstrained}.)

We choose a diagonal Gaussian for its efficiency and analytic entropy.
This choice may call to mind the Laplace approximation
technique, where a second-order Taylor expansion around the
maximum-a-posteriori estimate gives a Gaussian approximation to the
posterior.  However, using a
Gaussian variational approximation is not equivalent to the Laplace
approximation \citep{opper2009variational}. Our approach is distinct
in another way: the posterior approximation in the original latent variable
space (Figure \ref{sub:A}) is non-Gaussian, because of the
inverse transformation $T^{-1}$ and its Jacobian.

\subsection{Automatic Differentiation for Stochastic Optimization}

We now seek to maximize the \gls{ELBO} in real coordinate space,
\begin{align}
  \mbmu^*,{\mbsigma^2}^*
  &=
  \argmax_{\mbmu,\mbsigma^2} \cL(\mbmu,\mbsigma^2)
  \quad\text{such that}\quad
  \mbsigma^2 \succ 0.
  \label{eq:elbo_unconstrained}
\end{align}
We can use gradient ascent to reach a local maximum of the \gls{ELBO}.
Unfortunately, we cannot apply automatic differentiation to the \gls
{ELBO} in this form. This is because the expectation defines an intractable
integral that depends on $\mbmu$ and $\mbsigma^2$; we cannot
directly represent it as a computer program. Moreover, the variance vector
$\mbsigma^2$ must remain positive.
Thus, we employ one final transformation: elliptical standardization%
\footnote{Also known as
a ``co-ordinate transformation'' \citep{rezende2014stochastic},
an ``invertible transformation'' \citep{titsias2014doubly},
and
the ``re-parameterization trick'' \citep{kingma2013auto}.}
\citep{hardle2012applied}, shown in Figures \ref{sub:B} and \ref{sub:C}.

First, re-parameterize the Gaussian distribution with the log of the
standard deviation, $\mbomega = \log(\mbsigma)$, applied
element-wise. The support of $\mbomega$ is now the real coordinate
space and $\mbsigma$ is always positive.  Then, define the
standardization
$\mbeta = S_{\mbmu,\mbomega}(\mbzeta) =
\diag(\exp(\mbomega^{-1})) (\mbzeta - \mbmu)$.
The standardization encapsulates the variational parameters; in return
it gives a fixed variational density \begin{align*} q (\mbeta \,;\,
  \mb{0}, \mbI) &= \cN
  (\mbeta \,;\, \mb{0}, \mbI)
  =
  \prod_{k=1}^K
  \cN
  (\eta_k \,;\, 0, 1).
\end{align*}
The standardization transforms the variational problem from
Equation \ref{eq:elbo_unconstrained} into
\begin{align*}
\mbmu^*,{\mbomega}^*
  &=
  \argmax_{\mbmu,\mbomega} \cL (\mbmu, \mbomega)\\
  &=
  \argmax_{\mbmu,\mbomega}
  \E_{\cN(\mbeta\,;\, \mb{0}, \mbI)}
  \bigg[
  \log p
  (\mbX,T^{-1}(S_{\mbmu,\mbomega}^{-1}(\mbeta)))
  +
  \log \big| \det J_{T^{-1}}(S_{\mbmu,\mbomega}^{-1}(\mbeta)) \big|
  \bigg] + \sum_{k=1}^K \omega_k,
\end{align*}
where we drop independent term from the calculation.
The expectation is now in terms of the standard Gaussian, and both parameters
$\mbmu$ and $\mbomega$ are unconstrained. (Figure \ref{sub:C}.) We push
the gradient inside the expectations and apply the chain rule to get
\begin{align}
  \nabla_\mbmu \cL
  &=
  \E_{\cN(\mbeta)}
  \left[
  \nabla_\mbtheta \log p(\mbX,\mbtheta)
  \nabla_{\mbzeta} T^{-1}(\mbzeta)
  +
  \nabla_{\mbzeta}
  \log \big| \det J_{T^{-1}}(\mbzeta) \big|
  \right],
  \label{eq:elbo_us_grad_mu}
  \\
  \nabla_{\omega_k} \cL
  &=
  \E_{\cN(\eta_k)}
  \left[
  \left(
  \nabla_{\theta_k} \log p(\mbX,\mbtheta)
  \nabla_{\zeta_k} T^{-1}(\mbzeta)
  +
  \nabla_{\zeta_k}
  \log \big| \det J_{T^{-1}}(\mbzeta) \big|
  \right)
  \eta_k
  \exp(\omega_k)
  \right]
  + 1.
  \label{eq:elbo_us_grad_L}
\end{align}
(Derivations in Appendix \ref{app:gradient_elbo}.)

We can now compute the gradients inside the expectation with automatic
differentiation.  This leaves only the expectation.  \gls {MC}
integration provides a simple approximation: draw $M$ samples
from the standard Gaussian and evaluate the empirical mean of the gradients
within the expectation \citep{robert1999monte}. This gives unbiased noisy
estimates of gradients of the \gls{ELBO}.

\subsection{Scalable Automatic Variational Inference}

Equipped with unbiased noisy gradients of the \gls{ELBO}, \gls{ADVI} implements
stochastic gradient ascent. (Algorithm \ref{alg:ADVI}.) We ensure convergence
by choosing a decreasing step-size schedule. In practice, we
use an adaptive schedule \citep{duchi2011adaptive} with finite memory.
(See Appendix \ref{app:adaGrad} for details.)

\gls{ADVI} has complexity $\mathcal{O}(2NMK)$ per iteration, where $M$ is the
number of \gls{MC} samples (typically between 1 and 10). Coordinate ascent
\gls{VI} has complexity $\mathcal{O}(2NK)$ per pass over the dataset.
We scale  \gls{ADVI} to large datasets using stochastic optimization
\citep{hoffman2013stochastic,titsias2014doubly}. The adjustment to
Algorithm \ref{alg:ADVI} is simple:
sample a minibatch of size $B \ll N$ from the dataset and scale the likelihood
of the model by $N/B$ \citep{hoffman2013stochastic}.
The stochastic extension of \gls{ADVI} has a per-iteration complexity $\mathcal
{O}(2BMK)$.

\begin{algorithm}[t]
  \caption{Automatic Differentiation Variational Inference}
  \SetAlgoLined
  \DontPrintSemicolon
  \BlankLine
  \KwIn{Dataset $\mbX=\mbx_{1:N}$, model $p(\mbX,\mbtheta)$.}
  Set iteration counter $i = 0$ and choose a stepsize sequence $\mbrho^{(i)}$.\;
  Initialize $\mbmu^{(0)} = \mb{0}$ and $\mbomega^{(0)} = \mb{0}$.\;
  \BlankLine
  \While{change in \gls{ELBO} is above some threshold}{
    Draw $M$ samples $\mbeta_m \sim \cN(\mb{0},\mbI)$ from
    the standard multivariate Gaussian.\;
    \BlankLine
    Invert the standardization
    $\mbzeta_m = \diag(\exp(\mbomega^{(i)}))\mbeta_m + \mbmu^ {
    (i)}$.\;
    \BlankLine
    Approximate $\nabla_\mbmu \cL$ and $\nabla_{\mbomega} \cL$
    using \gls{MC} integration (Equations \ref{eq:elbo_us_grad_mu}
    and \ref{eq:elbo_us_grad_L}).\; \BlankLine
    Update
    $\mbmu^{(i+1)}            \longleftarrow
    \mbmu^{(i)} + \mbrho^{(i)}\nabla_\mbmu \cL
    \text{ and }\:
    \mbomega^{(i+1)}  \longleftarrow
    \mbomega^{(i)}  + \mbrho^{(i)}\nabla_{\mbomega} \cL$.\;
    \BlankLine
    Increment iteration counter.\;
  }
  Return
  $\mbmu^*             \longleftarrow \mbmu^{(i)}
  \text{ and }\:
  \mbomega^*    \longleftarrow \mbomega^{(i)}$.\;
  \label{alg:ADVI}
\end{algorithm}

\section{Empirical Study}
\label{sec:empirical}

We now study \gls{ADVI} across a variety of models. We compare its speed and
accuracy to two \gls{MCMC} sampling
algorithms: \gls{HMC} \citep{girolami2011riemann} and the \gls{NUTS}%
\footnote{\gls{NUTS} is an adaptive extension of \gls{HMC}. It is the default
sampler in Stan.} \citep
{hoffman2014nuts}.
We assess \gls{ADVI} convergence by tracking the \gls{ELBO}; assessing
convergence with \gls{MCMC} techniques is less straightforward. To place
\gls{ADVI}
and \gls{MCMC} on a common scale, we report predictive accuracy on held-out
data as a function of time. We approximate the Bayesian posterior
predictive using \gls{MC} integration. For the \gls{MCMC} techniques, we plug in
posterior samples into the likelihood. For \gls{ADVI}, we do the
same by drawing a sample from the posterior approximation at fixed intervals
during the optimization.
We initialize \gls{ADVI} with a draw from a standard Gaussian.

We explore two hierarchical regression models, two matrix
factorization models, and a mixture model. All of these models have
nonconjugate prior structures.
We conclude by analyzing a dataset of $250\,000$ images, where we report
results across a range of minibatch sizes $B$.

\subsection{A Comparison to Sampling: Hierarchical Regression Models}

We begin with two nonconjugate regression models: linear regression with
\gls{ARD} \citep{bishop2006pattern} and hierarchical logistic regression
\citep{gelman2006data}.

\parhead{Linear Regression with \gls{ARD}.} This is a sparse linear
regression model with a hierarchical prior structure. (Details in Appendix
\ref{app:linreg_ard}.) We simulate a dataset with $250$ regressors such that
half of
the regressors have no predictive power. We use $10\,000$ training samples and
hold out $1000$ for testing.

\parhead{Logistic Regression with Spatial Hierarchical Prior.} This is a
hierarchical logistic regression model from political science. The
prior captures dependencies, such as states and regions, in
a polling dataset from the United States 1988 presidential election
\citep{gelman2006data}.
The model is nonconjugate and would require
some form of approximation to derive a \gls{VI} algorithm. (Details in
Appendix \ref{app:logreg}.)

We train using $10\,000$ data point and withhold $1536$ for evaluation.
The regressors contain age, education, and state and region indicators.
The dimension of the regression problem is $145$.

\begin{figure}[!t]
\centering
  \begin{subfigure}[b]{2.6in}
  \hspace*{-0.25in}
    \includegraphics[width=2.4in]{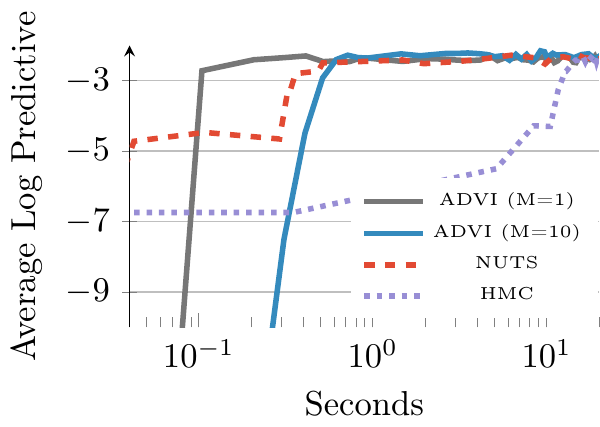}
    \caption{Linear Regression with \gls{ARD}}
    \label{sub:linreg}
  \end{subfigure}
  \begin{subfigure}[b]{2.6in}
    \includegraphics[width=2.6in]{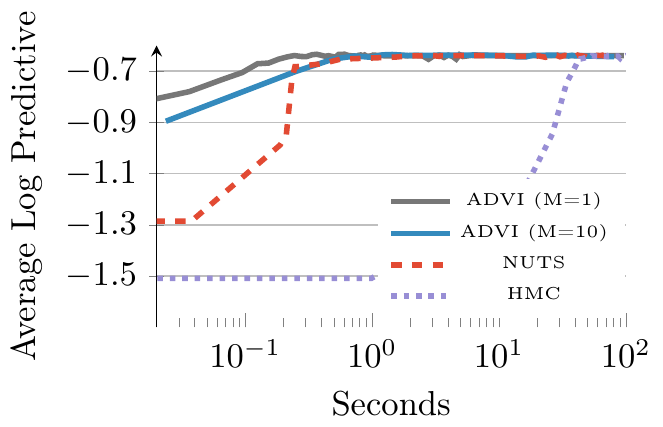}
    \vspace*{-0.15in}
    \caption{Hierarchical Logistic Regression}
    \label{sub:election88}
  \end{subfigure}
  \caption{Hierarchical Generalized Linear Models.}
  \label{fig:regression_models}
\end{figure}

\parhead{Results.}
Figure \ref{fig:regression_models} plots average log predictive accuracy as a
function of time. For these simple models, all methods reach the same predictive
accuracy. We study \gls{ADVI} with two settings of $M$, the number of \gls{MC}
samples used to estimate gradients. A single sample per iteration is
sufficient; it also is the fastest. (We set $M=1$ from here on.)

\subsection{Exploring nonconjugate Models: Non-negative Matrix Factorization}
\label{sub:NMF}

\begin{figure}[t]
\centering
  \begin{subfigure}[t]{2.6in}
    \includegraphics[width=2.6in]{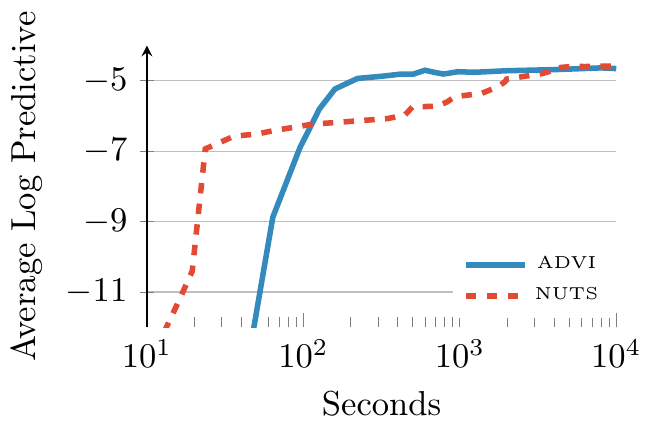}
    \vspace*{-0.20in}
    \caption{Gamma Poisson Predictive Likelihood}
    \vspace*{0.1in}
    \label{sub:mf_gap_pred}
  \end{subfigure}
  \begin{subfigure}[t]{2.6in}
    \includegraphics[width=2.6in]{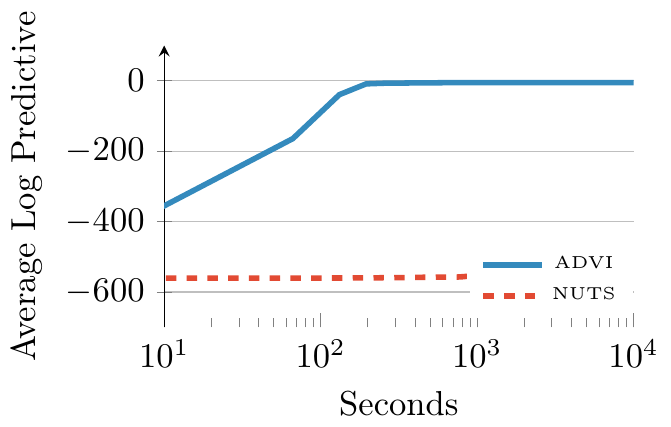}
    \vspace*{-0.20in}
    \caption{Dirichlet Exponential Predictive Likelihood}
    \vspace*{0.1in}
    \label{sub:mf_dirichlet_exp_pred}
  \end{subfigure}

  \begin{subfigure}[b]{2.6in}
  \centering
    \includegraphics[width=1.6in]{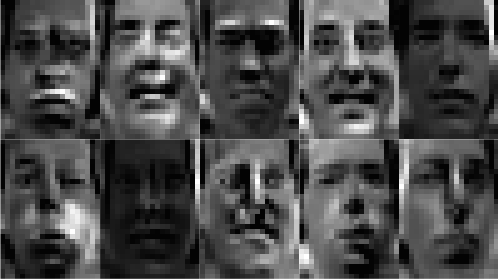}
    \caption{Gamma Poisson Factors}
    \label{sub:mf_gap_factors}
  \end{subfigure}
  \begin{subfigure}[b]{2.6in}
  \centering
    \includegraphics[width=1.6in]{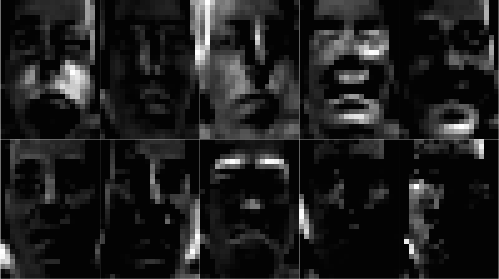}
    \caption{Dirichlet Exponential Factors}
    \label{sub:mf_dirichlet_exp_factors}
  \end{subfigure}
  \caption{Non-negative matrix factorization of the Frey Faces dataset.}
  \label{fig:nmf}
\end{figure}

We continue by exploring two nonconjugate non-negative matrix factorization
models: a constrained Gamma Poisson model
\citep{canny2004gap} and a Dirichlet Exponential model.
Here, we show how easy it is to explore new models using \gls{ADVI}. In both
models, we use the Frey Face dataset, which contains $1956$
frames ($28\times20$ pixels) of facial expressions extracted from a video
sequence.

\parhead{Constrained Gamma Poisson.}
This is a Gamma Poisson factorization model with an ordering constraint: each
row of the Gamma matrix goes from small to large values. (Details in Appendix
\ref{app:gap}.)

\parhead{Dirichlet Exponential.}
This is a nonconjugate Dirichlet Exponential factorization model with a Poisson
likelihood. (Details in Appendix \ref{app:dir_exp}.)

\parhead{Results.}
Figure \ref{fig:nmf} shows average log predictive accuracy as well as ten
factors recovered from both models.
\gls{ADVI} provides an order of magnitude speed improvement over \gls{NUTS}
(Figure \ref{sub:mf_gap_pred}). \gls{NUTS} struggles with the Dirichlet
Exponential
model (Figure \ref{sub:mf_dirichlet_exp_pred}). In both cases, \gls{HMC} does not
produce any useful samples within a budget of one hour; we omit \gls{HMC} from
the plots.

The Gamma Poisson model (Figure \ref{sub:mf_gap_factors}) appears to pick
significant frames out of the dataset. The Dirichlet Exponential factors
(Figure \ref{sub:mf_dirichlet_exp_factors}) are sparse and indicate
components of the face that move, such as eyebrows, cheeks, and the mouth.

\subsection{Scaling to Large Datasets: Gaussian Mixture Model}
\label{sub:scaling_gmm}

We conclude with the \gls{GMM} example we highlighted earlier. This is a
nonconjugate \gls {GMM} applied to color image histograms. We place a Dirichlet
prior on the mixture proportions, a
Gaussian prior on the component means, and a lognormal prior on the standard
deviations. (Details in Appendix \ref{app:gmm}.) We explore the image\textsc
{clef}
dataset, which has $250\,000$ images \citep{villegas13_CLEF}. We withhold
$10\,000$ images for evaluation.

In Figure \ref{sub:gmm_1000} we randomly select $1000$ images and train a model
with $10$
mixture components. \gls{NUTS} struggles to find an adequate
solution and \gls{HMC} fails altogether. This is likely due to label switching,
which can affect \gls{HMC}-based techniques in mixture models
\citep{stan-manual:2015}.

Figure \ref{sub:gmm_adsvi} shows \gls{ADVI} results on the full dataset. Here we
use
\gls{ADVI} with stochastic subsampling of minibatches from the dataset
\citep{hoffman2013stochastic}. We increase the number of mixture components to
$30$. With a minibatch size of $500$ or larger, \gls{ADVI} reaches high
predictive accuracy. Smaller minibatch sizes lead to suboptimal solutions, an
effect also observed in \citep {hoffman2013stochastic}. \gls{ADVI} converges
in about two hours.

\glsresetall{}
\section{Conclusion}

We develop \gls{ADVI} in Stan. \gls{ADVI} leverages automatic transformations,
an implicit non-Gaussian variational approximation, and automatic differentiation.
This is a valuable tool. We can explore many
models, and analyze large datasets with ease. We emphasize that \gls {ADVI} is
currently available as part of Stan; it is ready for anyone to use.

\subsubsection*{Acknowledgments}

We acknowledge our amazing colleagues and funding sources here.

\clearpage
\appendix

\section{Transformation of the Evidence Lower Bound}
\label{app:elbo_unconstrained}

Recall that $\mbzeta = T(\mbtheta)$ and that the variational approximation in
the real coordinate space is
$q(\mbzeta \,;\, \mbmu, \mbsigma^2)$.

We begin with the \gls{ELBO} in the original latent variable space. We then
transform the latent variable space of to the real coordinate space.
\begin{align*}
  \cL
  &=
  \int
  q(\mbtheta\,;\,\mbphi)
  \log \left[\frac{p(\mbX,\mbtheta)}{q(\mbtheta\,;\,\mbphi)}\right]
  \dif\mbtheta
  \\
  &=
  \int
  q(\mbzeta \,;\, \mbmu, \mbsigma^2)
  \log
  \left[
  \frac
  {p (\mbX,T^{-1}(\mbzeta)) \big| \det J_{T^{-1}}(\mbzeta) \big|}
  {q(\mbzeta \,;\, \mbmu, \mbsigma^2)}
  \right]
  \dif \mbzeta
  \\
  &=
  \int q(\mbzeta \,;\, \mbmu, \mbsigma^2)
  \log
  \left[
  p (\mbX,T^{-1}(\mbzeta)) \big| \det J_{T^{-1}}(\mbzeta) \big|
  \right]
  \dif \mbzeta
  -
  \int q(\mbzeta \,;\, \mbmu, \mbsigma^2)
  \log
  \left[
  q(\mbzeta \,;\, \mbmu, \mbsigma^2)
  \right]
  \dif \mbzeta
  \\
  &= \E_{q(\mbzeta)} \left[ \log p (\mbX,T^{-1}(\mbzeta))
     +
     \log \big| \det J_{T^{-1}}(\mbzeta) \big| \right]
     -
     \E_{q(\mbzeta)} \left[ \log q(\mbzeta \,;\, \mbmu, \mbsigma^2) \right]
\end{align*}
The variational approximation in the real coordinate space is a Gaussian.
Plugging in its entropy gives the \gls{ELBO} in the real coordinate space
\begin{align*}
\cL
&=
\E_{q(\mbzeta) } \left[ \log p (\mbX,T^{-1}(\mbzeta))
+
\log \big| \det J_{T^{-1}}(\mbzeta) \big| \right]
+
\frac{1}{2}K\left(1+\log(2\pi)\right)
+
\sum_{k=1}^K \log \sigma_k.
\end{align*}

\section{Gradients of the Evidence Lower Bound}
\label{app:gradient_elbo}

First, consider the gradient with respect to the $\mbmu$ parameter of the
standardization. We exchange the order of the gradient and the integration
through the dominated convergence theorem \citep{cinlar2011probability}. The
rest is the chain rule for differentiation.
\begin{align*}
  \nabla_\mbmu \cL
  &=
  \nabla_\mbmu
  \Big\{
  \E_{\cN(\mbeta\,;\, \mb{0}, \mbI)}
  \left[
  \log p
  (\mbX,T^{-1}(S_{\mbmu,\mbomega}^{-1}(\mbeta)))
  +
  \log \big| \det J_{T^{-1}}(S_{\mbmu,\mbomega}^{-1}(\mbeta)) \big|
  \right]\\
  &\qquad\quad+ \frac{K}{2}(1+\log(2\pi)) + \sum_{k=1}^K \log \sigma_k
  \Big\}\\
  &=
  \E_{\cN(\mbeta\,;\, \mb{0}, \mbI)}
  \left[
  \nabla_\mbmu
  \left\{
  \log p
  (\mbX,T^{-1}(S^{-1}(\mbeta))
  +
  \log \big| \det J_{T^{-1}}(S^{-1}(\mbeta)) \big|
  \right\}
  \right] \\
  &=
  \E_{\cN(\mbeta\,;\, \mb{0}, \mbI)}
  \left[
  \nabla_\mbtheta \log p(\mbX,\mbtheta)
  \nabla_{\mbzeta} T^{-1}(\mbzeta)
  \nabla_{\mbmu} S_{\mbmu,\mbomega}^{-1} (\mbeta)
  +
  \nabla_{\mbzeta}
  \log \big| \det J_{T^{-1}}(\mbzeta) \big|
  \nabla_{\mbmu} S_{\mbmu,\mbomega}^{-1} (\mbeta)
  \right] \\
  &=
  \E_{\cN(\mbeta\,;\, \mb{0}, \mbI)}
  \left[
  \nabla_\mbtheta \log p(\mbX,\mbtheta)
  \nabla_{\mbzeta} T^{-1}(\mbzeta)
  +
  \nabla_{\mbzeta}
  \log \big| \det J_{T^{-1}}(\mbzeta) \big|
  \right]
\end{align*}

Similarly, consider the gradient with respect to the $\mbomega$ parameter of the
standardization. The gradient with respect to a single component,
$\omega_k$, has a clean form. We abuse the $\nabla$ notation to maintain
consistency with the rest of the text (instead of switching to $\partial$).
\begin{align*}
  \nabla_{\omega_k} \cL
  &=
  \nabla_{\omega_k}
  \Big\{
  \E_{\cN(\mbeta\,;\, \mb{0}, \mbI)}
  \left[
  \log p
  (\mbX,T^{-1}(S_{\mbmu,\mbomega}^{-1}(\mbeta))
  +
  \log \big| \det J_{T^{-1}}(S_{\mbmu,\mbomega}^{-1}(\mbeta)) \big|
  \right]\\
  &\qquad\quad+ \frac{K}{2}(1+\log(2\pi)) + \sum_{k=1}^K \log( \exp
  (\omega_k))
  \Big\}\\
  &=
  \E_{\cN(\eta_k)}
  \left[
  \nabla_{\omega_k}
  \big\{
  \log p
  (\mbX,T^{-1}(S_{\mbmu,\mbomega}^{-1}(\mbeta)))
  +
  \log \big| \det J_{T^{-1}}(S_{\mbmu,\mbomega}^{-1}(\mbeta)) \big|
  \big\}
  \right] + 1\\
  &=
  \E_{\cN(\eta_k)}
  \left[
  \left(
  \nabla_{\theta_k} \log p(\mbX,\mbtheta)
  \nabla_{\zeta_k} T^{-1}(\mbzeta)
  +
  \nabla_{\zeta_k}
  \log \big| \det J_{T^{-1}}(\mbzeta) \big|
  \right)
  \nabla_{\omega_k} S_{\mbmu,\mbomega}^{-1}(\mbeta))
  \right]
  + 1.\\
  &=
  \E_{\cN(\eta_k)}
  \left[
  \left(
  \nabla_{\theta_k} \log p(\mbX,\mbtheta)
  \nabla_{\zeta_k} T^{-1}(\mbzeta)
  +
  \nabla_{\zeta_k}
  \log \big| \det J_{T^{-1}}(\mbzeta) \big|
  \right)
  \eta_k
  \exp(\omega_k)
  \right]
  + 1.
\end{align*}

\clearpage
\section{Running \gls{ADVI} in Stan}
\label{app:stan}

Use \texttt{git} to checkout the \texttt{feature/bbvb} branch from
\texttt{https://github.com/stan-dev/stan}. Follow instructions to build Stan.
Then download \texttt{cmdStan} from \texttt
{https://github.com/stan-dev/cmdstan}. Follow instructions to build \texttt
{cmdStan} and
compile your model. You are then ready to run \gls{ADVI}.

The syntax is

\begin{figure}[!h]
\centering
\ttfamily
\begin{tabular}{rl}
./myModel & experimental variational\\
          & grad\_samples=M $\qquad\qquad\qquad$( $M = 1$ default )\\
          & data file=myData.data.R\\
          & output file=output\_advi.csv\\
          & diagnostic\_file=elbo\_advi.csv
\end{tabular}
\end{figure}

where \texttt{myData.data.R} is the dataset in the \texttt{R} language \texttt
{dump} format. \texttt{output\_advi.csv} contains samples from the posterior and
\texttt{elbo\_advi.csv} reports the \gls{ELBO}.

\section{Transformations of Continuous Probability Densities}
\label{app:jacobian}

We present a brief summary of transformations, largely based on \citep
{olive2014statistical}.

Consider a univariate (scalar) random variable $X$ with probability density
function $f_X(x)$. Let $\mathcal{X} = \supp(f_X(x))$ be the support of
$X$. Now consider another random variable $Y$ defined as $Y = T(X)$.
Let $\mathcal{Y} = \supp(f_Y(y))$ be the support of $Y$.

If $T$ is a one-to-one and differentiable function from $\mathcal{X}$ to
$\mathcal{Y}$, then $Y$ has probability density function
\begin{align*}
  f_Y(y)
  &=
  f_X(T^{-1}(y))
  \left|
  \frac{\dif T^{-1}(y)}{\dif y}
  \right|.
\end{align*}
Let us sketch a proof. Consider the cumulative density function
$Y$. If the transformation $T$ is increasing, we directly apply its
inverse to the cdf of $Y$. If the transformation $T$ is decreasing, we apply its
inverse to one minus the cdf of $Y$. The probability density function is the
derivative of the cumulative density function. These things combined give the
absolute value of the derivative above.

The extension to multivariate variables $\mb{X}$ and $\mb{Y}$ requires a
multivariate version of the absolute value of the derivative of the inverse
transformation. This is the the absolute determinant of the Jacobian,
$|\det J_{T^{-1}}(\mb{Y})|$ where the Jacobian is
\begin{align*}
  J_{T^{-1}}(\mb{Y})
  &=
  \left(
  \begin{matrix}
    \frac{\partial T_1^{-1}}{\partial y_1}
    &
    \cdots
    &
    \frac{\partial T_1^{-1}}{\partial y_K}\\
    \vdots & & \vdots\\
    \frac{\partial T_K^{-1}}{\partial y_1}
    &
    \cdots
    &
    \frac{\partial T_K^{-1}}{\partial y_K}\\
  \end{matrix}
  \right).
\end{align*}

Intuitively, the Jacobian describes how a transformation warps unit volumes
across spaces. This matters for transformations of random variables, since
probability density functions must always integrate to one. If the
transformation is linear, then we can drop the Jacobian adjustment; it evaluates
to one. Similarly, affine transformations, like elliptical standardizations, do
not require Jacobian adjustments; they preserve unit volumes.

\section{Setting a Stepsize Sequence for \gls{ADVI}}
\label{app:adaGrad}

We use adaGrad \citep{duchi2011adaptive} to adaptively set the stepsize sequence
in \gls{ADVI}. While adaGrad offers attractive convergence properties, in
practice it can be slow because it has infinite memory. (It tracks the norm of
the gradient starting from the beginning of the optimization.) In \gls{ADVI} we
randomly initialize the variational approximation, which can be far from the
true posterior. This makes adaGrad take very small steps for the rest of the
optimization, thus slowing convergence. Limiting adaGrad's memory speeds up
convergence in practice, an effect also observed in training neural networks
\citep{rmsprop}. (See \cite{kingma2014adam} for an analysis of these trade-offs
and a method that combines benefits from both.)

Consider the stepsize $\mbrho^{(i)}$ and a gradient vector $\mb{g}^{(i)}$ at
iteration $i$. The $k$th element of $\mbrho^{(i)}$ is
\begin{align*}
  \rho_k^{(i)}
  &=
  \frac{\eta}
  {\tau + \sqrt{s^{(i)}_k}}
\end{align*}
where, in adaGrad, $\mb{s}$ is the gradient vector squared, summed over all
times steps since the start of the optimization. Instead, we limit this
to the past ten iterations and compute $\mb{s}$ as
\begin{align*}
  s^{(i)}_k
  &=
  {g^2_k}^{(i-10)} + {g^2_k}^{(i-9)} + \cdots + {g^2_k}^{(i)}.
\end{align*}
(In practice, we implement this recursively to save memory.) We set
$\eta=0.1$ and $\tau=1$ as the default values we use in Stan.

\section{Linear Regression with Automatic Relevance Determination}
\label{app:linreg_ard}

Linear regression with \gls{ARD} is a high-dimensional sparse
regression model \citep{bishop2006pattern,drugowitsch2013variational}. We
describe the model below. Stan code is in Figure \ref{fig:code_linreg}.

The inputs are $\mbX=\mbx_{1:N}$ where each $\mbx_n$ is $D$-dimensional. The
outputs are $\mb{y}=y_{1:N}$ where each $y_n$ is $1$-dimensional. The weights
vector $\mb{w}$ is
$D$-dimensional. The likelihood
\begin{align*}
  p(\mb{y} \mid \mbX, \mb{w}, \tau)
  &=
  \prod_{n=1}^N \cN \left( y_n \mid \mb{w}^\top \mbx_n \;,\; \tau^{-1}\right)
\end{align*}
describes measurements corrupted by iid Gaussian noise with unknown variance
$\tau^{-1}$.

The \gls{ARD} prior and hyper-prior structure is as follows
\begin{align*}
  p(\mb{w},\tau,\mb{\alpha})
  &=
  p(\mb{w},\tau \mid \mb{\alpha})
  p(\mb{\alpha})\\
  &=
  \cN \left( \mb{w} \mid 0 \,,\, (\tau\diag[\mb{\alpha}])^{-1} \right)
  \Gam (\tau \mid a_0, b_0)
  \prod_{i=1}^D \Gam({\alpha}_i \mid c_0, d_0)
\end{align*}
where $\mb{\alpha}$ is a $D$-dimensional hyper-prior on the weights, where each
component gets its own independent Gamma prior.

We simulate data such that only half the regressions have predictive power.
The results in Figure \ref{sub:linreg} use $a_0 = b_0 = c_0 = d_0 = 1$ as
hyper-parameters for the Gamma priors.

\section{Hierarchical Logistic Regression}
\label{app:logreg}

Hierarchical logistic regression models dependencies in an intuitive and
powerful way. We study a model of voting preferences from the 1988 United
States presidential election. Chapter 14.1 of \citep{gelman2006data}
motivates the model and explains the dataset. We also describe the
model below.
Stan code is in Figure \ref{fig:code_election88}, based on \citep
{stan-manual:2015}.
\begin{align*}
  \Pr(y_n=1)
  &=
  \sigma
  \bigg(
  \beta^0
  + \beta^\text{female}\cdot\text{female}_n
  + \beta^\text{black}\cdot\text{black}_n
  + \beta^\text{female.black}\cdot\text{female.black}_n \\
  &\qquad\quad + \alpha^\text{age}_{k[n]}
  + \alpha^\text{edu}_{l[n]}
  + \alpha^\text{age.edu}_{k[n],l[n]}
  + \alpha^\text{state}_{j[n]}
  \bigg)\\
  \alpha^\text{state}_j
  &\sim
  \cN
  \left(
  \alpha^\text{region}_{m[j]}
  + \beta^\text{v.prev}\cdot\text{v.prev}_j
  \,,\, \sigma^2_\text{state}
  \right)
\end{align*}
where $\sigma(\cdot)$ is the sigmoid function (also know as the logistic
function).

The hierarchical variables are
\begin{align*}
  \alpha^\text{age}_k
  &\sim
  \cN \left(0\,,\, \sigma^2_\text{age} \right)
  \text{ for } k = 1,\ldots,K\\
  \alpha^\text{edu}_l
  &\sim
  \cN \left(0\,,\, \sigma^2_\text{edu} \right)
  \text{ for } l = 1,\ldots,L\\
  \alpha^\text{age.edu}_{k,l}
  &\sim
  \cN \left(0\,,\, \sigma^2_\text{age.edu} \right)
  \text{ for } k = 1,\ldots,K, l = 1,\ldots,L\\
  \alpha^\text{region}_m
  &\sim
  \cN \left(0\,,\, \sigma^2_\text{region} \right)
  \text{ for } m = 1,\ldots,M.
\end{align*}

The variance terms all have uniform hyper-priors, constrained between 0 and 100.

\section{Non-negative Matrix Factorization: Constrained Gamma Poisson Model}
\label{app:gap}

The Gamma Poisson factorization model is a powerful way to analyze
discrete data matrices \citep{canny2004gap,cemgil2009bayesian}.

Consider a $U \times I$ matrix of observations. We find it helpful
to think of $u=\{1,\cdots,U\}$ as users and $i=\{1,\cdots,I\}$ as items, as in a
recommendation system
setting. The generative process for a Gamma Poisson model with $K$ factors is
\begin{enumerate}
  \item For each user $u$ in $\{1,\cdots,U\}$:
  \begin{itemize}
    \item For each component $k$, draw $\theta_{uk} \sim \Gam(a_0, b_0)$.
  \end{itemize}
  \item For each item $i$ in $\{1,\cdots,I\}$:
  \begin{itemize}
    \item For each component $k$, draw $\beta_{ik} \sim \Gam(c_0, d_0)$.
  \end{itemize}
  \item For each user and item:
  \begin{itemize}
    \item Draw the observation $y_{ui} \sim
    \text{Poisson}(\mbtheta_u^\top\mbbeta_i)$.
  \end{itemize}
\end{enumerate}

A potential downfall of this model is that it is not uniquely identifiable:
scaling $\mbtheta_u$ by $\alpha$ and $\mbbeta_i$ by $\alpha^{-1}$ gives the same
likelihood. One way to contend with this is to constrain either vector to be a
positive, ordered vector during inference. We constrain each $\mbtheta_u$
vector in our model in this fashion. Stan code is in Figure \ref
{fig:code_nmf_pf}. We set $K=10$ and all the Gamma hyper-parameters to 1 in our
experiments.

\section{Non-negative Matrix Factorization: Dirichlet Exponential Model}
\label{app:dir_exp}

Another model for discrete data is a Dirichlet Exponential model. The
Dirichlet enforces uniqueness while the exponential promotes sparsity. This
is a non-conjugate model that does not appear to have been studied in the
literature.

The generative process for a Dirichlet Exponential model with $K$ factors is
\begin{enumerate}
  \item For each user $u$ in $\{1,\cdots,U\}$:
  \begin{itemize}
    \item Draw the $K$-vector $\mbtheta_{u} \sim \text{Dir}(\mb{\alpha}_0)$.
  \end{itemize}
  \item For each item $i$ in $\{1,\cdots,I\}$:
  \begin{itemize}
    \item For each component $k$, draw $\beta_{ik} \sim \text{Exponential}
    (\lambda_0)$. \end{itemize}
  \item For each user and item:
  \begin{itemize}
    \item Draw the observation $y_{ui} \sim
    \text{Poisson}(\mbtheta_u^\top\mbbeta_i)$.
  \end{itemize}
\end{enumerate}

Stan code is in Figure \ref{fig:code_nmf_dir_exp}. We set $K=10$, $\alpha_0 = 1000$
for each component, and $\lambda_0 = 0.1$. With this configuration of
hyper-parameters, the
factors $\mbbeta_i$ are sparse and appear interpretable.

\section{Gaussian Mixture Model}
\label{app:gmm}

The \gls{GMM} is a powerful probability model. We use it to group a dataset of
natural images based on their color histograms. We build a high-dimensional \gls
{GMM} with a Gaussian prior for the mixture means, a lognormal prior for the
mixture standard deviations, and a Dirichlet prior for the mixture components.

The images are in $\mbX = \mbx_{1:N}$ where each $\mbx_n$ is $D$-dimensional and
there are $N$ observations. The likelihood for the images is
\begin{align*}
  p(\mbX \mid \mbtheta, \mbmu, \mbsigma)
  &=
  \prod_{n=1}^N
  \prod_{k=1}^K
  \theta_k
  \prod_{d=1}^D
  \cN(x_{nd}\mid\mu_{kd},\sigma_{kd})
\end{align*}
with a Dirichlet prior for the mixture proportions
\begin{align*}
  p(\mbtheta)
  &=
  \text{Dir}(\mbtheta\,;\, \mb{\alpha}_0),
\end{align*}
a Gaussian prior for the mixture means
\begin{align*}
  p(\mbmu)
  &=
  \prod_{k=1}^D
  \prod_{d=1}^D
  \cN(\mu_{kd}\,;\, 0, \sigma_\mu)
\end{align*}
and a lognormal prior for the mixture standard deviations
\begin{align*}
  p(\mbsigma)
  &=
  \prod_{k=1}^D
  \prod_{d=1}^D
  \text{logNormal}(\sigma_{kd}\,;\, 0, \sigma_\sigma)
\end{align*}

The dimension of the color histograms in the image\textsc{clef} dataset is $D =
576$. These a concatenation of three $192$-length histograms, one for each
color channel (red, green, blue) of the images.

We scale the image histograms to have zero mean and unit variance and set
$\alpha_0 = 10\,000$, $\sigma_\mu=0.1$ and $\sigma_\mu$. \gls{ADVI} code is in
Figure \ref{fig:code_gmm_diag}. The stochastic data subsampling version of the code is
in Figure \ref{fig:code_gmm_diag_adsvi}.

\begin{figure}[htbp]
\centering
  \parbox[c]{3.75in}{\input{stan_programs/linreg.tex}}
  \caption{Stan code for Linear Regression with Automatic Relevance
  Determination.}
  \label{fig:code_linreg}
\end{figure}

\begin{figure}[htbp]
\centering
  \parbox[c]{3.75in}{\input{stan_programs/election88.tex}}
  \caption{Stan code for Hierarchical Logistic Regression,
  from \citep{stan-manual:2015}.}
  \label{fig:code_election88}
\end{figure}

\begin{figure}[htbp]
\centering
  \parbox[c]{3.75in}{\input{stan_programs/nmf_pf.tex}}
  \caption{Stan code for Gamma Poisson non-negative matrix factorization model.}
  \label{fig:code_nmf_pf}
\end{figure}

\begin{figure}[htbp]
\centering
  \parbox[c]{3.95in}{\input{stan_programs/nmf_dir_exp.tex}}
  \caption{Stan code for Dirichlet Exponential non-negative matrix factorization
  model.}
  \label{fig:code_nmf_dir_exp}
\end{figure}

\begin{figure}[htbp]
\centering
  \parbox[c]{5.0in}{\input{stan_programs/gmm_diag.tex}}
  \caption{\gls{ADVI} Stan code for the \gls{GMM} example.}
  \label{fig:code_gmm_diag}
\end{figure}

\begin{figure}[htbp]
\centering
  \parbox[c]{5.00in}{\input{stan_programs/gmm_diag_adsvi.tex}}
  \caption{\gls{ADVI} Stan code for the \gls{GMM} example, with stochastic
  subsampling of the dataset.}
  \label{fig:code_gmm_diag_adsvi}
\end{figure}

\clearpage
\bibliographystyle{unsrtnat}
\bibliography{advi.bib}

\end{document}